\documentclass{article}

\usepackage{microtype}
\usepackage{booktabs}
\usepackage{graphicx}
\usepackage{caption}
\usepackage{subcaption}
\usepackage{amsmath}
\usepackage{hyperref}

\usepackage[accepted]{icml2018}

\icmltitlerunning{Goal-oriented Trajectories for Efficient Exploration}

\begin{document}

\twocolumn[
\icmltitle{Goal-oriented Trajectories for Efficient Exploration} 

\icmlsetsymbol{equal}{*}
\begin{icmlauthorlist}
\icmlauthor{Fabio Pardo}{ril}
\icmlauthor{Vitaly Levdik}{ril}
\icmlauthor{Petar Kormushev}{ril}
\end{icmlauthorlist}
\icmlaffiliation{ril}{Robot Intelligence Lab, Imperial College London, UK}
\icmlcorrespondingauthor{Fabio Pardo}{f.pardo@imperial.ac.uk}
\icmlkeywords{Reinforcement Learning, Exploration}

\vskip 0.3in
]

\printAffiliationsAndNotice{}

\begin{abstract}

Exploration is a difficult challenge in reinforcement learning and even recent state-of-the art curiosity-based methods rely on the simple epsilon-greedy strategy to generate novelty. We argue that pure random walks do not succeed to properly expand the exploration area in most environments and propose to replace single random action choices by random goals selection followed by several steps in their direction. This approach is compatible with any curiosity-based exploration and off-policy reinforcement learning agents and generates longer and safer trajectories than individual random actions. To illustrate this, we present a task-independent agent that learns to reach coordinates in screen frames and demonstrate its ability to explore with the game Super Mario Bros. improving significantly the score of a baseline DQN agent.

\end{abstract}

\section{Introduction}

Exploration is an inherent part of a reinforcement learning algorithm allowing an agent to learn about its environment. The goal of exploration is to reduce uncertainty about the environment's transitions and to discover rewards. The most common approach to exploration in absence of any knowledge about the environment is to perform random actions. As knowledge is gained, the agent can use it to attempt to increase its performance by taking greedy actions, while retaining some chance to choose random actions to further explore the environment ($\varepsilon$-greedy exploration).

\begin{figure}
    \centering
    \begin{subfigure}{.48\linewidth}
        \centering
        \includegraphics[width=\textwidth]{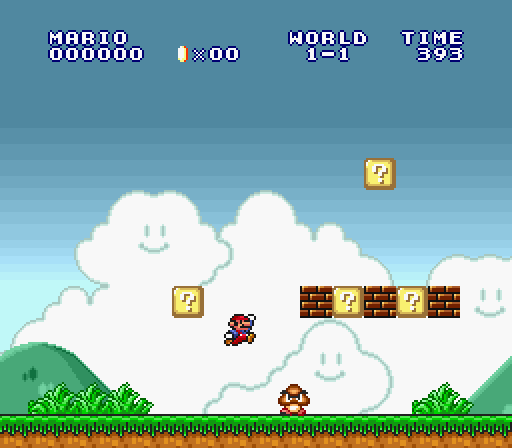}
    \end{subfigure}
    \hfill
    \begin{subfigure}{.48\linewidth}
        \centering
        \includegraphics[width=\textwidth]{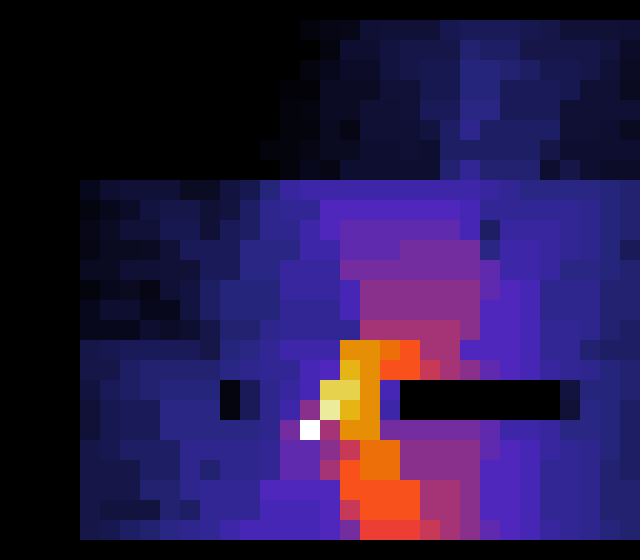}
    \end{subfigure}
    \caption{Discovering how to navigate Super Mario Bros. (All Stars) by reaching random screen coordinates. Left: a capture of the screen. Right: a ground-truth Q-map frame showing the expected distance to screen coordinates.
    }
    \label{fig:fig1}
\end{figure}

However, if rewards are sparse or are not sufficiently informative to allow performance improvements, $\varepsilon$-greedy fails to explore sufficiently far. Several methods have been described that bias the agent's actions towards novelty mostly by using optimistic initialization or curiosity signals \cite{oudeyer2007intrinsic, oudeyer2009intrinsic, schmidhuber1991possibility, schmidhuber2010formal}. While the first one is mainly compatible with tabular methods, the second relies on intrinsic rewards motivating, for example, information gain \cite{kearns1999efficient, brafman2002r}, state visitation counts \cite{bellemare2016unifying, tang2017exploration} or prediction error \cite{stadie2015incentivizing, pathak2017curiosity}. While these methods show success in sparse reward environments, they dynamically modify the agent's reward function and thus make the environment's MDP appear non-stationary to the agent, while still mostly relying on completely random actions to discover novelty.

The main issue with random actions is that they might waste a lot of time by cancelling each other or push into obstacles and can frequently lead the agent to terminal states. We argue that if during random exploration, actions are still chosen in a coherent way it is possible to extend the boundaries of the exploration area much faster. To achieve this purpose, we propose to replace most of the random actions with short trajectories towards random goals.

As a practical implementation, we propose to use screen coordinates as goals and present a novel and efficient algorithm, that we call Q-map, in charge of learning to reach them directly from screen frames.

\section{Q-maps}

\begin{figure}
    \centering
    \includegraphics[width=\linewidth]{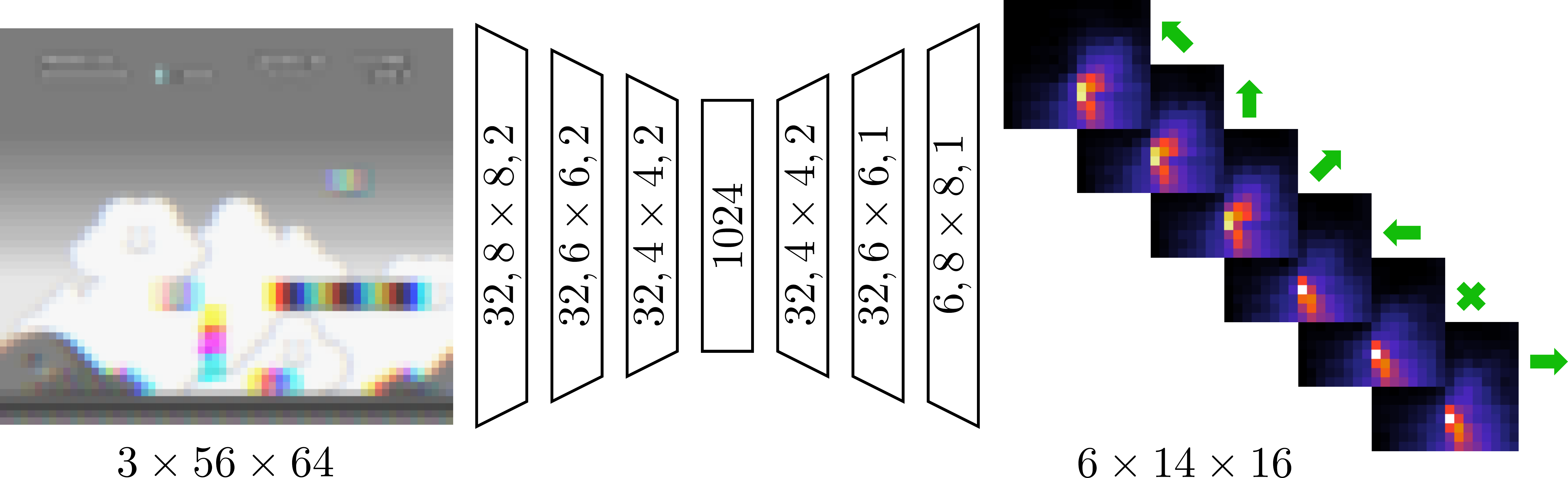}
    \caption{Q-map neural network architecture. A stack of grey-scale frames as input, with $3$ convolutions, $1$ hidden layer and $3$ transposed convolutions to output a stack of Q-maps, one for every possible action choice}
    \label{fig:architecture}
\end{figure}

The core concept of the proposed Q-map algorithm is that a single transition $(s, a, s')$ allows us to update, off-policy, the estimated distance from $s$ to any goal $g$ when taking action $a$. In \cite{andrychowicz2017hindsight} goals used for such updates are taken from the set of states visited later in experienced trajectories. While this allows to focus on goals that can effectively be visited later from a given state, it lacks the ability to learn to reach other goals never experienced in following states. On the contrary we propose to use all possible goals for each update. This is performed by only providing frames to an artificial neural network and requesting the 3D tensors of Q-values $Q(s,a,g)$ for all actions and goals at once. Other works focused on neural networks taking states and goals in input as proposed for Universal Value Function Approximators \cite{schaul2015universal}.

Q-map agent uses screen coordinates as goals, where the learned Q-values represent a notion of the expected distance to a point in screen space. As these goals are likely correlated and depend on visual cues, we expect that an auto-encoder-like architecture would be able to efficiently tackle large output matrix. Convolutions are applied to the input frames followed by fully connected layers and deconvolutions (convolution transpose) to output a set of 2D frames (Q-maps), one for each action. In each frame the rows and columns directly indicate a coordinate in screen space, covering the entire visible screen.

The approach to provide only the state as the input, expecting the output to consist of values for all goals is similar to the implementation by \cite{hasselt2010double}, where DQN network produces Q values for every action, given a state in input. For DQN agent, the approach allows for efficient choice of greedy action with a single pass through the network. In the case of the Q-map agent, similar implementation, illustrated in Figure \ref{fig:architecture}, allows the network to exploit locality, to converge faster and generalize visual cues, to update towards all possible goals at once and to efficiently select goals.

\begin{algorithm}[tb]
    \caption{Training a Q-map network $\mathcal{Q}$}
    \label{alg:q-map-update}
    \begin{algorithmic}
        \FOR {\text{iteration} = 1, 2, ...}
            \STATE Sample a batch $b$ of transitions from a replay buffer $\mathcal{B}$
            \FOR {\text each transition $(s, a, s', x', y', \text{term})^{(i)}$ in $b$}
                \IF {$\text{term} = \text{True}$}
                    \STATE $Q^{(i)} \gets 0$
                \ELSE
                    \STATE $Q^{(i)} \gets \gamma \text{ clip}(\max_{a'} \mathcal{Q}(s')[:,:,a'], 0, 1)$
                    \STATE $Q^{(i)}[y',x'] \gets 1$
                \ENDIF
            \ENDFOR
            \STATE Update $\mathcal{Q}$ to reduce $\sum_i \left\| Q^{(i)} - \mathcal{Q}(s^{(i)})[:,:,a^{(i)}] \right\|^2_2$
        \ENDFOR
    \end{algorithmic}
\end{algorithm}

Given a transition $(s,a,s',x',y')$, the value $Q(s,a,g)$ for any coordinates $g$ can be updated using a variant of Q-learning, with the following target:
\begin{equation}
    y =
    \begin{cases}
        1                           & \text{if } (x', y') = g \\
        0                           & \text{else if } s' \text{ is terminal} \\
        \gamma \max_{a'} Q(s',a',g) & \text{otherwise}
    \end{cases} 
\end{equation}
Where $\gamma$ is a discount factor used for the Q-map, not necessarily the same as the one discounting the rewards from the environment. This is efficiently done using the 3D tensors of Q-values from the network as detailed in Algorithm \ref{alg:q-map-update}.

Q-maps are suited for environments in which it is possible to locate the agent's position in screen coordinates, which could either be provided by the environment (e.g. from the RAM in video games), or a separate classifier could be trained to localise the agent in the environment. While coordinates are used to create the target for the Q-learning, it does not however preclude one from only using raw frames as input for the Q-map agent. In some games, such as Super Mario Bros. (used later in the experiments), the screen scrolls while the player's avatar moves, and thus only a portion of an entire level is shown at once. In the proposed Q-map implementation we chose to use the coordinates available on the screen as possible goals and not coordinates over an entire level, thus the map is local to the area around the agent.

While distance to goals could more directly be represented by the expected number of steps, the decay factor forces values to be bounded, with a value of $0$ for points wich are unreachable and $1$ for points immediately reachable. The Q-values naturally decay for points which are never reached. However, it is important to note that Q-map agent is not a planner, the Q-values represent an expected value which can become quite unreliable after several time steps, especially as the agent lacks prediction to deal with dynamic environments. Nevertheless we expect the agent to have sufficiently accurate local estimations, which can further be improved by requesting a new Q-map every time step towards a chosen goal.

\section{Combining Q-maps with reinforcement learning}

\begin{algorithm}[tb]
    \caption{Action selection with a Q-map DQN agent}
    \label{alg:ac-select}
    \begin{algorithmic}
    \IF {$\text{rand}() < \Pr(\text{random action})$}
        \STATE $a \gets \text{random action}$
    \ELSE
        \IF {$g$ was reached or is not in screen}
            \STATE $T \gets 0$
        \ENDIF
        \IF {$T = 0$}
            \IF {$\text{rand}() < \Pr(\text{random goal})$}
            \STATE $G \gets$ coordinates of Q-map values in range
                \IF {$G \neq \emptyset$}
                    \STATE $a \gets$ greedy DQN action
                    \STATE $B \gets$ coordinates of $G$ compatible with $a$
                    \IF {$\text{rand}() < \Pr(\text{biased goal})$ and $B \neq \emptyset$}
                        \STATE $g \gets$ random goal from $B$
                    \ELSE
                        \STATE $g \gets$ random goal from $G$
                        \STATE $a \gets$ greedy Q-map action 
                    \ENDIF
                    \STATE $T \gets$ expected time to $g$
                \ELSE
                    \STATE $a \gets$ random action
                \ENDIF
            \ELSE
                \STATE $a \gets $ greedy DQN action
            \ENDIF
        \ELSE
            \STATE $a \gets$ greedy Q-map action towards the goal
            \STATE $T \gets T-1$
        \ENDIF
    \ENDIF
    \STATE {\bfseries return} $a$
    \end{algorithmic}
\end{algorithm}

The proposed Q-map exploration agent can be used with any off-policy reinforcement learning agents and can learn simultaneously with it. A combined agent then consists of an exploration agent learning Q-maps of the environment and a task-learner agent learning to maximize the rewards in the environment. At every time step, if a goal has already been chosen, the Q-map agent provides the next action towards it and decreases the remaining time allowed to reach it, while if no goal exists, there is a probability to select a new one or to request an action from the task-learner agent. On top of these two agents it is also necessary to keep some amount of random actions in order to efficiently train both of them. The process is detailed in Algorithm \ref{alg:ac-select}.

The number of steps spent following goals is not accurately predictable thus ensuring that the average proportion of exploratory steps (random and goal-oriented actions) follows a scheduled proportion $\varepsilon$ is not straightforward. To achieve a good approximation we dynamically adjust the probability of selecting new goals to make a running average of the proportion of exploratory steps match the scheduled exploration. This allows us to compare the performance of our proposed agent and baseline DQN with a similar proportion of exploratory actions.

With $\varepsilon$-greedy, a fight can happen between the random actions and the greedy actions resulting in difficulty to properly move. With our proposed approach this phenomenon can be exacerbated even more since the exploratory steps are larger. To reduce this effect, with some fixed probability, a biasing of the goal selection is performed by first asking the task-learner agent for the next greedy action it would like to perform and keeping only goals compatible with that action.

\section{Experiments}

\begin{figure*}
    \centering
    \begin{subfigure}{\textwidth}
        \centering
        \includegraphics[width=\textwidth]{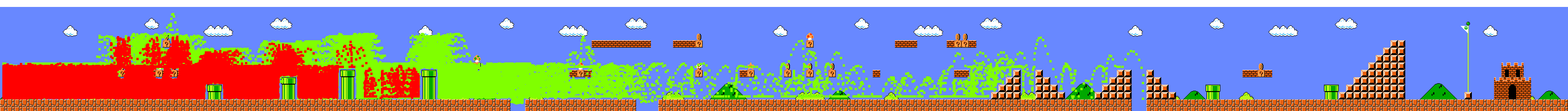}
    \end{subfigure}
    \hfill
    \begin{subfigure}{\textwidth}
        \centering
        \includegraphics[width=\textwidth]{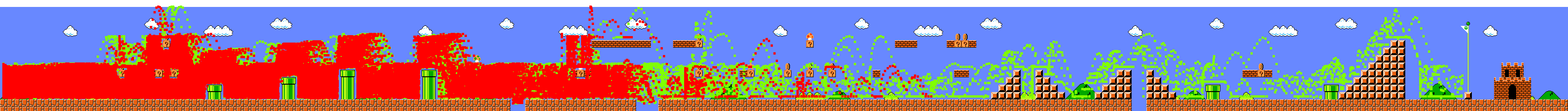}
    \end{subfigure}
    \caption{Binary visitation masks after 2M steps for \textit{Super Mario Bros. (All-Stars)} level 1. Top: Random walk (red) compared against Q-map random goal walk (green). Bottom: DQN $\varepsilon$-greedy (red) compared with Q-map DQN (green).}
    \label{fig:full_level_comparisons}
\end{figure*}

\subsection{Agent}

The proposed Q-map-DQN agent is composed of two DQN-based sub-agents \cite{mnih2015human}, one to learn the task-independent Q-maps and one to learn the task from environmental rewards. The implementation is based on OpenAI Baselines' DQN \cite{baselines}, using TensorFlow \cite{tensorflow2015-whitepaper}. Inputs are a stack of three $64\times45$ grayscale frames normalized ($-1$ for white and $1$ for black). Each sub-agent uses a separate artificial neural network but the same architecture is used for the encoder visible in Figure \ref{fig:architecture}. We used double Q-learning \cite{hasselt2010double} with target network updates every $1000$ steps, prioritized experience replay \cite{schaul2015prioritized} (with default parameters from Baselines) using a shared buffer of $500,000$ steps but separate priorities. The training starts after $1000$ steps and the networks are used after $2000$ steps. Both networks are trained every $4$ steps with independent batches of size $32$. Two Adam optimizers \cite{kingma2014adam} are used with learning rate $10^-4$ and default other hyperparameters from TensorFlow's implemetation. The probability of greedy actions from DQN increases from $0$ to $0.95$ linearly during the first $75\%$ steps.

The DQN sub-agent outputs $6$ Q-values, one for each action, uses dueling \cite{wang2015dueling}, relu activations and a discount factor of $0.99$.
The Q-map sub-agent outputs $6$ frames of size $16 \times 14$ ($4$ times lower resolution than the input), totalling $1344$ Q-values. The deconvolutions (convolutions transpose) use the same kernel shapes as in the encoder but smaller strides, and the elu activation function. Goals are selected within an expected range of $15$ to $30$ timesteps while the time given to reach them contains a $50\%$ supplement to account for possible random movements interfering with the trajectories. The goal selection is biased with a probability of $0.5$ to match the greedy action from the DQN sub-agent. The discount factor ($0.9$) decays the value towards impossible states to $0$ and force the agent to reach goals as quickly as possible. Finally, the probability of taking completely random action at any time is decayed from $0.1$ to $0.05$ during the first $75\%$ steps.

\subsection{Environment}

To measure the impact of the proposed exploration compared to epsilon greedy, we train both our agent and DQN (same agent deprived of the Q-map) on the first level of the game Super Mario Bros. (All Stars) from the Super Nintendo Entertainment System (SNES) console, using the Retro Learning Environment (RLE) framework \cite{bhonker2016playing} based on OpenAI Gym \cite{brockman2016openai}.
Transitions are deterministic, and the action set is limited to the $6$ most basic ones: no action, move left and right, jump up and diagonally to the left or to the right. Terminations by touching enemies or are detected from the RAM and only original SNES rewards are used and divided by $100$. No bonus was provided when moving to the right as it is originally implemented in RLE and no penalty for dying. Typical rewards are $0.5$ for breaking brick blocks, $1$ for killing enemies, $2$ for collecting coins, $4$ for throwing turtle shells, $10$ for eating consumables such as mushrooms, and $50$ for reaching the final flag. The coordinates of Mario and of the scrolling windows were extracted from the RAM. Episodes are naturally limited by the timer of $400$ seconds present in the game, which corresponds to $2394$ steps. Videos are available at: \url{https://sites.google.com/view/got-exploration}.

\subsection{Results}

\begin{figure}[t]
    \centering
    \includegraphics[width=\linewidth]{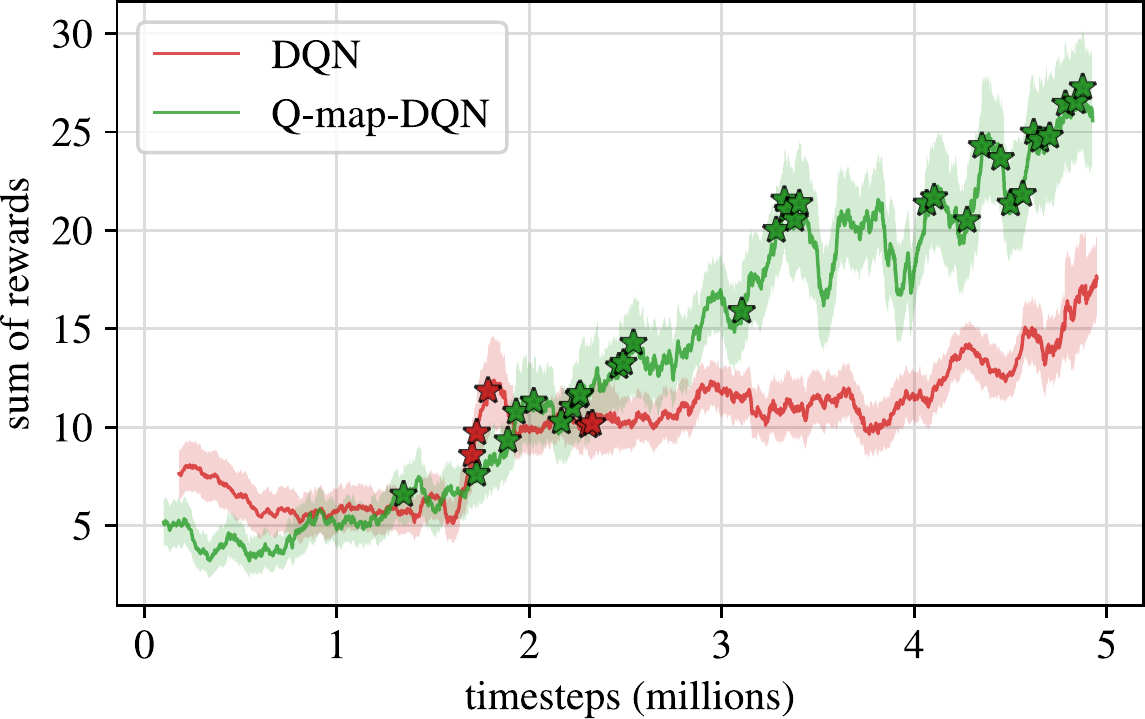}
    \caption{Performance comparison between a baseline $\varepsilon$-greedy DQN (red) and the proposed Q-map based DQN (green). Stars indicate timesteps at which the agents reached the flag (end of the level). The proposed agent significantly outperforms the baseline.}
    \label{fig:score}
\end{figure}

To verify that Q-map walk is more efficient than random walk in pure exploration, we first ran both types of walks, with no task-learner agent present, for $2$ million time steps and collected the coordinates traversed by Mario. Figure \ref{fig:full_level_comparisons} (top) shows that Q-map walk succeeds to push the boundaries of the exploration much further away than random walk and almost manages to reach the flag. To verify that Q-map effectively improves the exploration when training DQN, we then trained the full proposed agent for $2$ million time steps. Figure \ref{fig:full_level_comparisons} (bottom) shows that the combination of Q-map and DQN succeeds to ultimately push exploration to the end of the level by reaching the flag.

Finally, to measure the performance of the proposed agent in terms of sum of rewards collected per episode and number of flags reached we trained it for $5$ million time steps and reported the results in Figure \ref{fig:score}. Our proposed agent quickly outperforms the baseline DQN agent and reaches a consistent $60\%$ better performance during the final third of the training. The stars in the figure indicate that the flag has been reached. While DQN reaches the flag $5$ times, our agent reaches it $31$ times.

\section{Conclusion and future directions}

We have shown that exploring the environment by following a coherent sequence of steps towards random goals can successfully expand the exploration boundaries and thus improve reinforcement learning agents' performance when compared to common $\varepsilon$-greedy approaches. We have proposed and evaluated a practical implementation of such an exploration method combined with DQN and achieved significant performance gain over the baseline version.

Further work is needed to better evaluate the generalization properties of the proposed Q-map agent, such as by testing on a variety of other levels of the same game.

\section*{Acknowledgements}

The research presented in this paper has been supported by Dyson Technology Ltd. and Samsung.

\bibliography{main}
\bibliographystyle{icml2018}

\end{document}